\documentclass[runningheads]{llncs}

\usepackage{eccv}

\usepackage{eccvabbrv}

\usepackage{graphicx}
\usepackage{booktabs}

\usepackage[accsupp]{axessibility}  % Improves PDF readability for those with disabilities.

\usepackage{hyperref}

\usepackage{orcidlink}
\usepackage{multirow}
\usepackage{multicol}
\usepackage{pifont}
\definecolor{graytext}{RGB}{140,140,140}
\definecolor{greenhighlight}{RGB}{232,245,233}
\definecolor{lightgray}{RGB}{248,248,248}
\definecolor{improvered}{RGB}{200,30,30}
\definecolor{medgray}{gray}{0.82}
\newcommand{\imp}[1]{{\color{improvered}\scriptsize$^{\uparrow#1}$}}
\usepackage{tabularx}
\usepackage[most]{tcolorbox}
\usepackage[table]{xcolor}
\newcommand{\methodname}{FrameRepeat}
\newcommand{\blfootnote}[1]{
    \begingroup
    \renewcommand{\thefootnote}{}
    \renewcommand{\hypertarget}{}  % 临时禁用超链接
    \footnote{#1}
    \addtocounter{footnote}{-1}
    \endgroup
}

\begin{document}

% ---------------------------------------------------------------
% TODO REVIEW: Replace with your title
\title{When Thinking Hurts: Mitigating Visual Forgetting in Video Reasoning via Frame Repetition} 

% TODO REVIEW: If the paper title is too long for the running head, you can set
% an abbreviated paper title here. If not, comment out.
\titlerunning{FrameRepeat}

% TODO FINAL: Replace with your author list. 
% Include the authors' OCRID for the camera-ready version, if at all possible.
\author{Xiaokun Sun*\inst{1,2} \and
Yubo Wang*\inst{1,2} \and
Haoyu Cao\inst{1,2} \and 
Linli Xu$\dagger$\inst{1,2}}

% TODO FINAL: Replace with an abbreviated list of authors.
\authorrunning{X. Sun, Y. Wang et al.}
% First names are abbreviated in the running head.
% If there are more than two authors, 'et al.' is used.

% TODO FINAL: Replace with your institution list.
\institute{University of Science and Technology of China \and
State Key Laboratory of Cognitive Intelligence\\
\email{\{sunxiaokun2020, wyb123, caohaoyu\}@mail.ustc.edu.cn} \\
\email{linlixu@ustc.edu.cn}}

\maketitle
\blfootnote{* Equal contribution, $\dagger$ Corresponding author. }
\begin{abstract}
Recently, Multimodal Large Language Models (MLLMs) ha\-ve demonstrated significant potential in complex visual tasks through the integration of Chain-of-Thought (CoT) reasoning. 
However, in Video Question Answering, extended thinking processes do not consistently yield performance gains and may even lead to degradation due to ``visual anchor drifting'', where models increasingly rely on self-generated text, sidelining visual inputs and causing hallucinations.
While existing mitigations typically introduce specific mechanisms for the model to re-attend to visual inputs during inference, these approaches often incur prohibitive training costs and suffer from poor generalizability across different architectures.
% In this paper, we observe that reinforcing physical visual signals by repeating key video frames at the input stage can effectively guide the model to maintain focus on core visual evidence during reasoning. 
% To address this, we propose \methodname, a lightweight enhancement framework that reinforces key video frames during the input stage. By automating the identification of frames to reinforce, \methodname enables Video-LLMs to focus on critical visual signals during reasoning.
To address this, we propose \methodname, an automated enhancement framework which features a lightweight repeat scoring module that enables Video-LLMs to autonomously identify which frames should be reinforced. 
% The core objective of \methodname\ is to model the repeat gain of each frame.
% We propose a novel training strategy, Add-One-In (AOI), which measures the log-probability shift of the correct answer when a single frame is repeated, thereby providing frame-level supervision for learning question-aware importance ranking.
% We propose a novel training strategy, Add-One-In (AOI), which leverages the output probabilities of the MLLM to generate supervision signals representing repeat gain. These signals are used to train a frame scoring network, thereby guiding the frame repetition.
We introduce a novel training strategy, Add-One-In (AOI), that uses MLLM output probabilities to generate supervision signals representing repeat gain. This can be used to train a frame scoring network, which guides the frame repetition behavior.
% Extensive experimental results across multiple backbone models and diverse datasets demonstrate both the generalizability of \methodname\ and its effectiveness in strengthening critical visual signals during the reasoning process.
Experimental results across multiple models and datasets demonstrate that \methodname\ is both effective and generalizable in strengthening important visual cues during the reasoning process.

\end{abstract}

\section{Introduction}
\label{sec:intro}
Recently, multimodal large language models have demonstrated significant potential in complex visual understanding tasks by incorporating Chain-of-Thought (CoT) reasoning~\cite{wei2023chainofthoughtpromptingelicitsreasoning,wu2025visual,xing2025caprl,feng2025video,team2025kimi,jiang2025corvid}. 
However, existing studies indicate that in certain visual reasoning scenarios, as the reasoning chain grows longer, the model's attention toward visual modalities diminishes significantly~\cite{luo2025thinking,jian2025look,yang2025look}. 
The prediction process increasingly shifts toward relying on self-generated linguistic tokens, which makes the model more prone to hallucinations and reasoning drift. 
This phenomenon is particularly pronounced in video reasoning, where the inherent sparsity of temporal information means that critical evidence is often confined to a few sparse frames. 
The remaining frames, largely redundant or contextual, make it significantly harder for the model to maintain focus on salient information during reasoning.
Furthermore, as videos contain much more complex information, their CoT processes are often longer than those for static images, further exacerbating this issue.

\begin{figure*}[t]
\centering
\includegraphics[width=\linewidth]{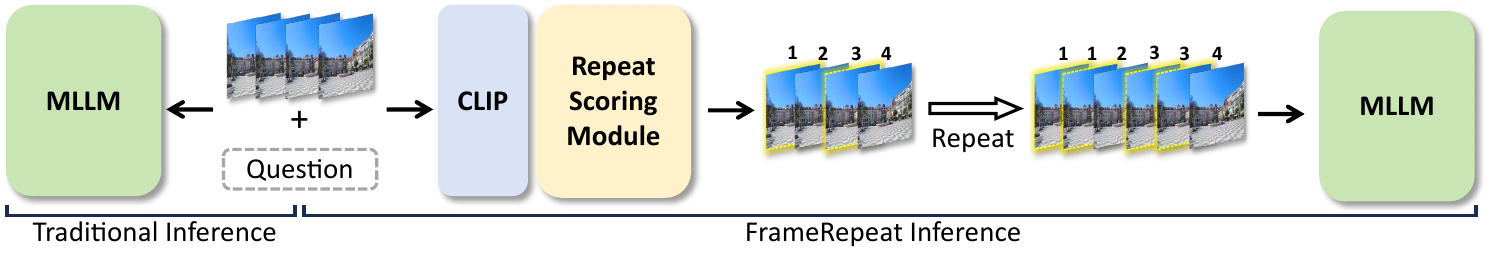}
\caption{\textbf{Comparison between traditional inference and FrameRepeat inference.}
In traditional inference, video frames are uniformly sampled and directly fed into the MLLM. In contrast, the proposed FrameRepeat first scores the frames during inference, selecting the frames with the highest scores, which are then repeated at their original positions and passed into the MLLM for enhanced understanding.}
\label{fig:intro}
\end{figure*}

While existing approaches typically mitigate this issue by introducing specific mechanisms—such as invoking external tools to re-incorporate pivotal visual information~\cite{li2025revisor,zheng2025deepeyes,hong2025deepeyesv2,he2025framethinker} or designing specialized loss functions to steer attention distribution~\cite{jian2025look}—they often entail computationally expensive training processes. 
In addition, the optimization gains achieved on a specific architecture frequently lack cross-model transferability, thereby limiting their broader application.

In contrast to the aforementioned methods, we argue that the root of this issue lies not in a deficiency of the model's reasoning capabilities, but rather in a signal imbalance between long-chain linguistic reasoning and visual input.
Specifically, if the essence of the problem is the dilution of visual signals during the reasoning process, then the primary objective should not be increasing reasoning steps, but rather maintaining the effective intensity of critical visual cues amidst extended CoT sequences.

We observe a simple yet consistent phenomenon: repeating specific video frames in the input stage enhances model performance without requiring any additional training. 
This finding suggests that increasing the redundancy of specific frames can, to some extent, counteract the dilution effect imposed by reasoning tokens. 
Based on this observation, we first analyze how frame repetition influences video reasoning. 
By examining the shifts in attention toward repeated segments during the thinking process, alongside the variations in the log probability of correct answers, we find that repeating key frames effectively bolsters the model's focus on these frames, and the impact of repeating different frames on the output log probability varies significantly.

Accordingly, we propose elevating frame repetition from a static heuristic to a reasoning-aware visual signal re-weighting process. 
Specifically, we introduce \methodname, an automated enhancement framework which features a lightweight repeat scoring module to predict an importance factor for each frame based on the frame's visual features and the question’s semantic representation, enabling adaptive repetition of pivotal frames at the input stage of Video-LLMs. 
We propose Add-One-In (AOI) to train \methodname: for each candidate frame, we measure the log-probability gain of the correct answer when that single frame is repeated, yielding a repeat gain signal that supervises the network to learn frame importance ranking.
During inference, \methodname\ selects the top-$k$ frames based on the repeat scoring module's output and repeats them at their original temporal positions. 
Notably, \methodname\ does not modify the backbone VLM architecture; instead, it strengthens the visual signal at the input level, ensuring the model maintains continuous focus on critical visual evidence throughout long-chain reasoning. Consequently, a repeat scoring module trained on one model can be seamlessly transferred to different architectures.

To comprehensively evaluate our approach, we conduct extensive experiments across two distinct backbone models, Qwen2.5-VL-Instruct-7B~\cite{bai2025qwen25vltechnicalreport}, Qwen3-VL-Thinking-8B~\cite{bai2025qwen3vltechnicalreport} and five representative benchmarks VideoMME~\cite{fu2025video}, Long\-Video\-Bench~\cite{wu2024longvideobench}, LVBench~\cite{wang2025lvbench}, MLVU~\cite{zhou2024mlvu}, Video-Holmes~\cite{cheng2025video}. 
The results demonstrate that \methodname\ significantly boosts model performance on video reasoning tasks while exhibiting strong generalization capabilities across different scenarios.

Our contributions can be summarized as follows:
\begin{itemize}
    \item We analyze the phenomenon of visual signal dilution in video reasoning scenarios, where long-chain reasoning leads to a decay in visual attention  and propose a novel approach that bolsters visual signal intensity through strategic frame repetition.

    \item We introduce \methodname, a lightweight cross-attention repeat scoring module that scores each frame's importance conditioned on the question, and selects the top-scoring frames for in-place duplication at the input stage. 
    To train this network, we propose Add-One-In (AOI), a marginal-contribution-based supervision strategy that measures each frame's value by the log-probability gain when it is individually repeated.

    \item Extensive experiments across five benchmarks and two distinct backbone models validate the effectiveness of \methodname\ in enhancing video reasoning performance and demonstrate its robust generalization capabilities.
\end{itemize}
\section{Related Work}
\label{sec:related_work}

\subsection{Reasoning in MLLMs}
Recent research has focused on enhancing the reasoning capabilities of Multimodal Large Language Models (MLLMs), inspired by the success of long chain-of-thought reasoning in language models~\cite{dong2025insight,yao2024mulberry,xiao2025fast,liu2025visual,cao2025ground,chen2025perception,wang2025internvl35advancingopensourcemultimodal}.
LLaVA-CoT~\cite{xu2025llava} introduces a structured four-stage reasoning pipeline—summary, caption, reasoning, and conclusion—with a stage-level beam search for effective inference-time scaling. 
R1-Onevision~\cite{yang2025r1} proposes a cross-modal formalization pipeline that converts visual inputs into formal textual representations and leverages reinforcement learning to develop generalized multimodal reasoning capabilities.

Despite these advances, recent studies have identified notable limitations in the CoT reasoning of current MLLMs. 
\cite{jian2025look} find that current visual reasoning models exhibit limited visual reflection, as their attention to visual information diminishes rapidly with longer generated responses. 
Similarly, \cite{yang2025look} points out that these models often excessively rely on textual information during the later stages of inference, neglecting the crucial integration of visual input. 
These findings highlight a critical challenge: how to prevent key visual signals from being diluted throughout the reasoning process, which is the central question explored in this work.

\subsection{Video Reasoning}
Extending reasoning capabilities to the video domain presents unique challenges, as it requires not only understanding individual frames but also modeling complex temporal dynamics, causal relationships, and event transitions across long sequences~\cite{fei2024video,ouyang2025spacer,wang2025videorft,chen2026omnivideo,ding2025videozoomer,wang2025fostering,liu2026videoauto}. 
Unlike image-based reasoning, video reasoning demands that models maintain a coherent understanding over time while integrating spatial and temporal cues for multi-step inference. 
Recent efforts have begun to explore chain-of-thought and reinforcement learning paradigms in this context. 
Specifically, VideoChat-R1~\cite{li2025videochat} enhances spatio-temporal perception in video reasoning through reinforcement fine-tuning. 
Video-R1~\cite{feng2025video} is the first to systematically explore the R1 paradigm for video reasoning via an upgraded T-GRPO algorithm, achieving strong results on spatial-temporal benchmarks.
However, in the context of video reasoning, introducing chain-of-thought often leads to performance degradation~\cite{luo2025thinking}. In this work, we analyze the underlying causes of this phenomenon and propose to mitigate it by enhancing the visual signal intensity of key frames.
\section{Dissecting the Effect of Frame Repetition}
\label{sec:pre_exper}
% In recent years, many works have introduced Chain-of-Thought (CoT) into video understanding, enabling models to reason about video-related questions in a step-by-step manner. 
% This approach not only allows models to better solve complex video question-answering tasks but also enhances the interpretability of their responses. 
% However, 
Prior works~\cite{luo2025thinking,jian2025look,yang2025look} have found that incorporating CoT may sometimes lead to a decline in model performance. 
This is because, as the chain of thought grows longer, the model tends to attend more to the generated text tokens while neglecting the visual information. 
This phenomenon is particularly severe in video reasoning, since the effective information in a video is often concentrated in a few key frames, while the remaining visual tokens provide background or redundant information, making it even harder for the model to attend to the critical information during reasoning. 
Given this observation, a natural question arises: \textbf{Is there a way to highlight the key information, so that critical visual signals can maintain their effective strength throughout the reasoning process?} 
We conduct several experiments to investigate this.

\begin{figure*}[t]
\centering
\includegraphics[width=1.0\linewidth]{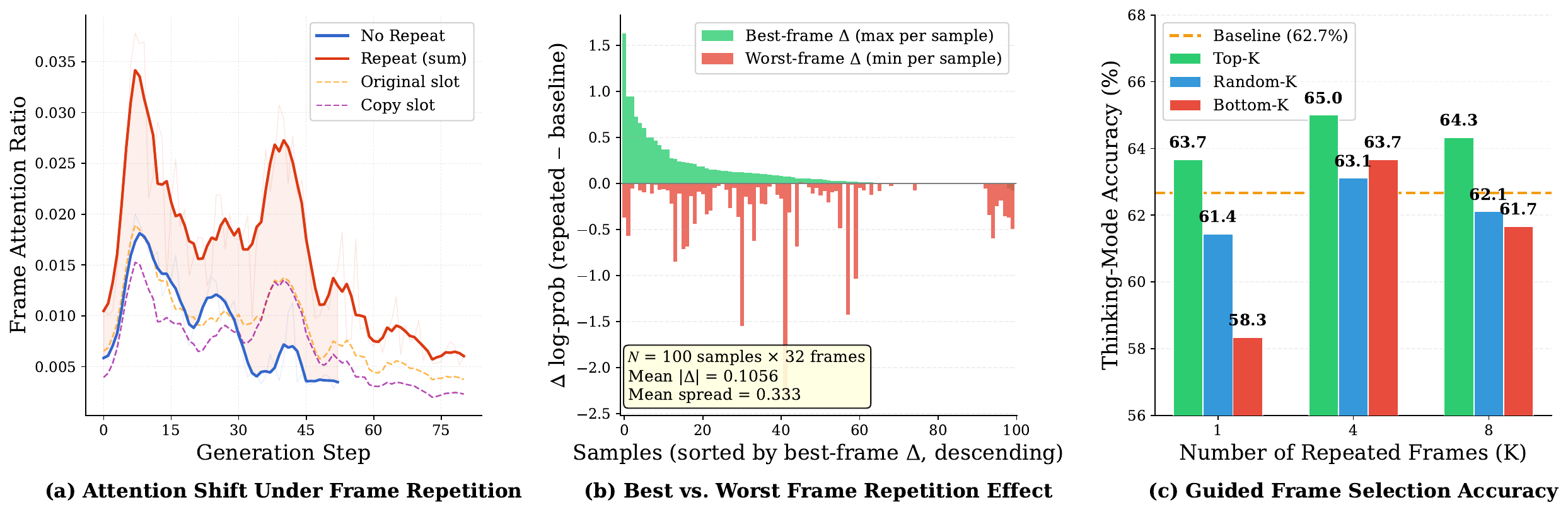}

\caption{\textbf{Empirical analysis of frame repetition.} 
\textbf{(a)} Attention ratio allocated to a repeated frame across generation steps. 
When a frame is repeated, it occupies two visual slots in the input sequence—the original slot and the copy slot—whose attention contributions are shown separately. 
Repetition significantly increases the model's overall attention to the target frame. 
\textbf{(b)} Per-sample best-frame (green) and worst-frame (red) $\Delta$log-prob, showing that repetition benefit is highly frame-dependent. 
\textbf{(c)} Thinking-mode accuracy when repeating Top-K, Random-K, or Bottom-K frames selected by direct-mode $\Delta$log-prob. Top-K consistently outperforms other strategies, validating the effectiveness of $\Delta$log-prob as a frame importance signal.}
\label{fig:pre_exper}

\end{figure*}

\subsection{Mitigating Visual Forgetting}
An intuitive idea is: since critical visual signals are gradually diluted during reasoning, could we enhance the presence of these key frames in the input—for example, by repeating them—to help the model sustain attention to critical information throughout long-chain reasoning? 
To verify this hypothesis, we design a simple experiment to observe the effect of frame repetition on attention distribution. 
Specifically, we use Qwen2.5-VL as the base model and randomly sample instances from VideoMME, with each video uniformly sampled into 128 frames as input.
We first perform one inference pass, identifying the key frame that receives the highest attention proportion from the answer token during answer generation, and record the attention ratio of each generated token to this key frame throughout the reasoning process. 
In the second pass, we duplicate this key frame and insert the copy immediately after its original position, re-running inference with 129 frames. 
We record the attention ratio curves over generation steps for both passes, where the attention under the repetition condition is the sum of the original and copied positions. 
The results are shown in~\cref{fig:pre_exper}(a). 
It can be clearly observed that without repetition, the model's attention to the key frame exhibits a sustained declining trend as generation progresses; with repetition, although the attention still decreases, it remains at a significantly higher level throughout the entire reasoning process.

\subsection{How Frame Repetition Affects Model Predictions}

% To verify that frame repetition has a significant impact on model outputs and that the effects of repeating different frames vary considerably, we conduct a per-frame log-probability scanning experiment on Qwen2.5-VL-7B. 
% Specifically, we randomly sample 100 instances from VideoMME, uniformly extract 128 frames per video as the baseline input, and let the model directly output the answer without chain-of-thought reasoning.
% For each sample, we uniformly select 32 frame positions and repeat them one at a time, measuring the change in log-probability of the correct answer ($\Delta$). 
% We plot the results in~\cref{fig:pre_exper}(b).
% The results show that repeating a single frame alone induces a mean $|\Delta\log\text{-prob}|$ of 0.1056, and in 94.0\% of samples, there exists at least one frame whose repetition increases the probability of the correct answer. 
% Furthermore, the influence varies significantly across frame positions within the same sample: the gap between the best and worst frame $\Delta$ averages 0.3325 (with a maximum of 2.3961), and in 60.0\% of samples this gap exceeds 0.1.
% These results demonstrate that frame repetition has a substantial impact on model outputs.

To evaluate the impact of frame repetition on model outputs, we conduct a per-frame log-probability scanning experiment on Qwen2.5-VL-7B. We randomly sample 100 instances from VideoMME, extract 128 frames per video as the baseline input, and let the model generate answers without chain-of-thought reasoning. 
For each sample, we repeat 32 randomly selected frames one at a time and measure the change in log-probability of the correct answer ($\Delta$). The results, shown in~\cref{fig:pre_exper}(b), reveal that repeating a single frame leads to a mean $|\Delta\log\text{-prob}|$ of 0.1056, and in 94.0\% of samples, at least one frame increases the probability of the correct answer. The influence varies widely across frame positions, with the average gap between the best and worst frame $\Delta$ being 0.3325 (maximum 2.3961), and in 60.0\% of samples, this gap exceeds 0.1. These findings highlight the significant impact of frame repetition on model outputs.

To validate whether the frame importance ranking obtained in direct (non-thinking) mode can effectively guide frame selection in thinking mode, and whether repeating multiple informative frames yields greater benefits than repeating a single frame, we design a Top-K vs.\ Random-K vs.\ Bottom-K experiment on 300 randomly sampled questions from VideoMME, using Qwen2.5-VL with 128 uniformly sampled base frames. 
For each sample, the experiment proceeds in two phases. In \textbf{Phase~1} (direct per-frame scan), we uniformly select 32 frame positions from the 128 base frames and repeat each one individually, measuring the change in log-probability of the ground-truth answer ($\Delta\text{log-prob}$) under direct mode. 
The 32 frames are then ranked by their $\Delta\text{log-prob}$ values. 
In \textbf{Phase~2} (thinking-mode evaluation), for each $K \in \{1, 4, 8\}$, we construct three conditions: (i)~\textbf{Top-K}, repeating the $K$ frames with the largest $\Delta\text{log-prob}$; (ii)~\textbf{Bottom-K}, repeating the $K$ frames with the smallest $\Delta\text{log-prob}$; and (iii)~\textbf{Random-K}, repeating $K$ randomly chosen frames (averaged over 3 independent trials to reduce variance). 
All conditions are evaluated in thinking mode with full chain-of-thought generation, and a no-repeat baseline is also measured.
We plot the results in~\cref{fig:pre_exper}(c).
Results confirm both hypotheses. 
The accuracy ordering Top-K $>$ Bottom-K and Top-K $>$ Random-K holds consistently across all $K$ values. 
This demonstrates that the lightweight direct-mode delta ranking is a reliable proxy for identifying frames that most benefit thinking-mode reasoning. Meanwhile, repeating multiple positively contributing frames amplifies the benefit beyond what any single frame can achieve: Top-4 reaches 65.0\%, a +2.3\% gain over the baseline, compared to +1.0\% for Top-1 (63.7\%), indicating that the effects of repeating individually informative frames accumulate rather than saturate at a single frame.

The above experimental findings suggest that if a method can enable the model to autonomously determine which frames to repeat, it would be possible to improve the model's performance on video reasoning tasks without introducing additional inference steps.
\section{Method}
\label{sec:methodology}
\begin{figure*}[t]
\centering
\includegraphics[width=\linewidth]{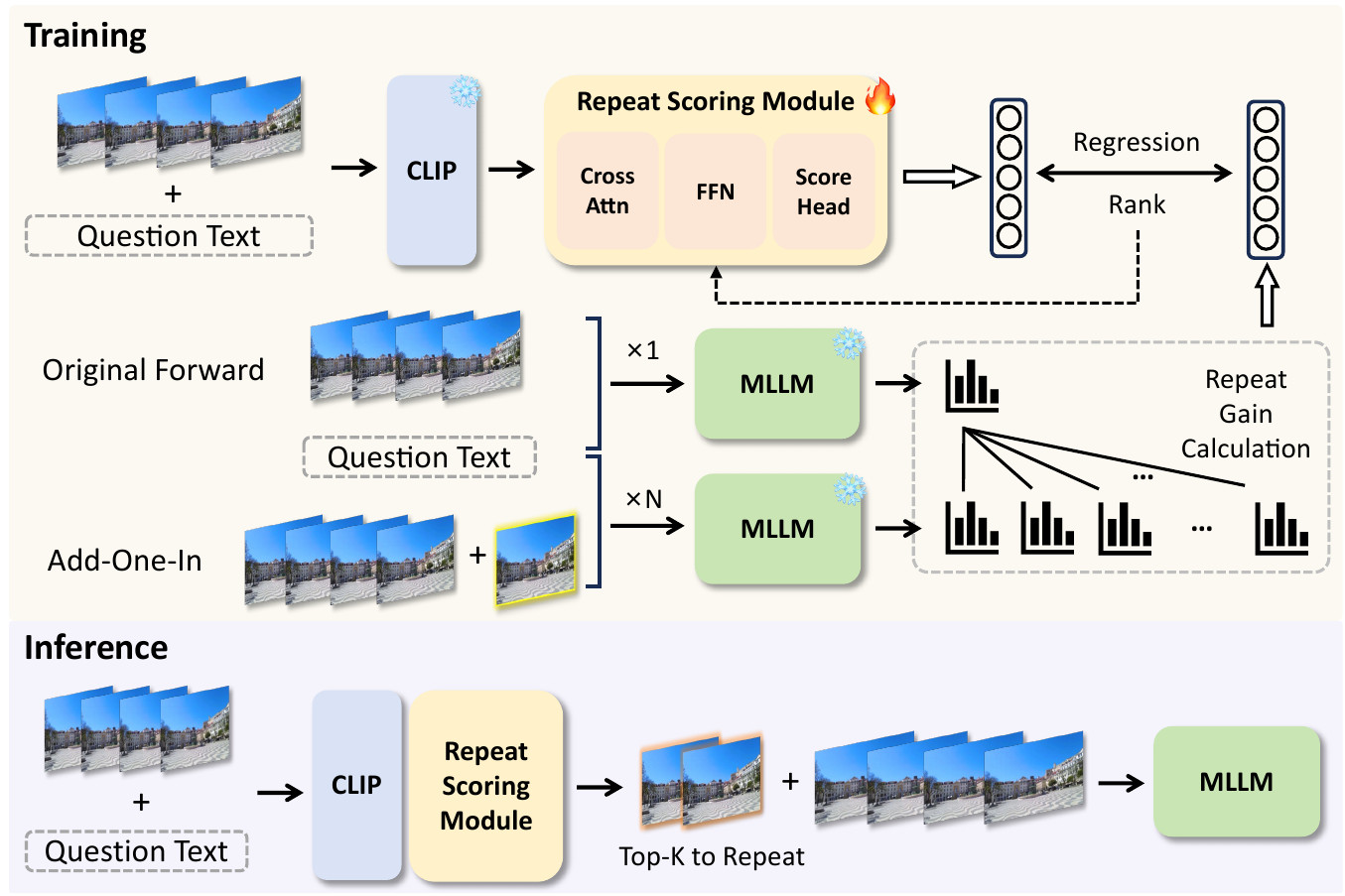}
\caption{\textbf{Overview of the FrameRepeat framework.}
% During training, we adopt an Add-One-In strategy to measure the output probability of the MLLM when each frame is individually repeated. Based on the resulting probability variations, we compute the repeat gain for each frame, which serves as supervision to train the repeat scoring module. During inference, the video frames and the question are first fed into the repeat scoring module to estimate the repetition importance of each frame. We then select the top-k most important frames for repetition and forward the augmented frame sequence to the MLLM to generate the final answer.
During training, we use the Add-One-In strategy to measure the MLLM output probability for each repeated frame. The resulting probability variations are used to compute the repeat gain, which trains the repeat scoring module. During inference, the video frames and question are passed through the scoring module to estimate the repetition importance of each frame. The top-k most important frames are then selected for repetition and fed into the MLLM for the final answer.
}
\label{fig:method}
\end{figure*}

As illustrated in~\cref{fig:method}, we propose \methodname, a method designed to enable Video-LLMs to autonomously learn which frames should be repeated during the input stage, thereby enhancing the signal strength of critical visual information during the reasoning process. 
In this section, we first introduce the architectural design of \methodname. 
Subsequently, we present Add-One-In (AOI), a frame-level value estimation strategy based on the repeat gain of repeating one frame. 
This strategy measures the extent to which repeating a specific frame contributes to the model's correct response, serving as the supervision signal for training the repeat scoring module. 
Finally, we detail the joint training objective, which comprises both regression and ranking losses.

\subsection{FrameRepeat Architecture}
During our experiments, we observed that critical frames are often highly correlated with the input query.
Therefore, when modeling the importance of each frame, it is necessary to incorporate textual information and enable interaction between text and visual modalities.

At the input stage, we first uniformly sample $N$ frames from the video as in previous methods, then encode each frame and the input question through CLIP~\cite{radford2021learning} to obtain their feature representations. 
Specifically, each frame is encoded as a $d$-dimensional vector $\mathbf{v}_i \in \mathbb{R}^d$ ($i = 1, \dots, N$), and the question text is encoded as a sequence of token embeddings $\mathbf{T} = [\mathbf{t}_1, \dots, \mathbf{t}_L] \in \mathbb{R}^{L \times d}$, where $L$ is the number of tokens. 
Since CLIP encodes each frame independently and is inherently agnostic to temporal ordering, we inject temporal information by adding sinusoidal positional encodings~\cite{vaswani2017attention}:

\begin{equation}
\tilde{\mathbf{v}}_i = \mathbf{v}_i + \text{PE}(i, N)
\label{eq:pos_enc}
\end{equation}

\noindent where $\text{PE}(i, N)$ denotes the sinusoidal positional encoding for the $i$-th frame in a sequence of length $N$.

To enable interaction between textual and visual features, we introduce a multi-head cross-attention module. 
Specifically, we use the visual features as queries and the text token sequence as keys and values:

\begin{equation}
\mathbf{h}_i = \text{CrossAttn}(\tilde{\mathbf{v}}_i, \mathbf{T}) = \text{softmax}\!\left(\frac{\mathbf{W}_Q \tilde{\mathbf{v}}_i \cdot (\mathbf{W}_K \mathbf{T})^\top}{\sqrt{d_h}}\right) \mathbf{W}_V \mathbf{T}
\label{eq:cross_attn}
\end{equation}

\noindent where $\mathbf{W}_Q, \mathbf{W}_K, \mathbf{W}_V \in \mathbb{R}^{d \times d}$ are learnable projection matrices and $d_h$ is the head dimension.
Each frame independently attends to the full question token sequence, yielding a question-conditioned frame representation via a residual connection followed by LayerNorm:

\begin{equation}
\hat{\mathbf{v}}_i = \text{LayerNorm}(\tilde{\mathbf{v}}_i + \mathbf{h}_i)
\label{eq:cross_residual}
\end{equation}

After obtaining the text-fused visual features, we further refine them through a two-layer feed-forward network (FFN) with a residual connection. 
Since the FFN operates on features that already encode question relevance from the cross-attention layer, it is able to capture non-linear patterns in the question-frame interaction --- for example, distinguishing frames that merely depict an entity mentioned in the question from those that capture the key action needed to answer it:

\begin{equation}
\mathbf{f}_i = \text{LayerNorm}(\hat{\mathbf{v}}_i + \text{FFN}(\hat{\mathbf{v}}_i))
\label{eq:ffn}
\end{equation}

We then map each refined frame feature to a scalar importance score through a two-layer projection head:

\begin{equation}
\hat{s}_i = \text{ScoreHead}(\mathbf{f}_i), \quad \text{ScoreHead}: \mathbb{R}^d \to \mathbb{R}^{d/4} \to \mathbb{R}^1
\label{eq:score_head}
\end{equation}

Finally, we add a zero-centered CLIP cosine similarity as a prior to the learned score:

\begin{equation}
s_i = \hat{s}_i + \lambda \cdot \left(\text{sim}(\mathbf{v}_i, \mathbf{t}) - \frac{1}{N}\sum_{j=1}^{N}\text{sim}(\mathbf{v}_j, \mathbf{t})\right)
\label{eq:clip_prior}
\end{equation}

\noindent where $\text{sim}(\cdot, \cdot)$ denotes the cosine similarity and $\lambda$ is a scaling coefficient. 
This provides a stable initialization signal for training, preventing the randomly initialized network from producing arbitrary rankings. 
The zero-centering ensures that the prior encodes only relative relevance across frames without shifting the absolute score level. 
During training, the MLLM backbone and CLIP encoder remain completely frozen, making the training process highly efficient and adding negligible overhead at inference time.

\subsection{Add-One-In}
The core task of the repeat scoring module is to assign an importance score to each frame such that frames with higher scores, when repeated, effectively improve the model's answer quality. 
This requires a training signal that directly reflects the impact of repeating a specific frame on the model's response. 
However, obtaining such frame-level supervision is non-trivial: existing video understanding datasets only provide question-answer pairs and lack annotations indicating which frames are critical for answering.

To address this, we propose the Add-One-In (AOI) training strategy. 
The key idea is to \textbf{automatically construct frame-level supervision by measuring the change in the model's log-probability of the correct answer after repeating a single frame}. 
Specifically, for each training sample, we first feed the original $N$ frames into the frozen MLLM and compute the baseline log-probability of the correct answer:

\begin{equation}
\ell_{\text{base}} = \log P\!\left(a^* \mid \{f_1, \dots, f_N\}, q\right)
\label{eq:baseline_logprob}
\end{equation}

\noindent where $a^*$ is the correct answer and $q$ is the question. 
Then, for each candidate frame $f_i$, we duplicate it once and insert the copy at its corresponding temporal position, forming an $(N\!+\!1)$-frame input, and recompute the log-probability:

\begin{equation}
\ell_i = \log P\!\left(a^* \mid \{f_1, \dots, f_i, f_i, \dots, f_N\}, q\right)
\label{eq:aoi_logprob}
\end{equation}

The difference between the two yields the \textit{Repeat Gain} of frame $f_i$:

\begin{equation}
\Delta_i = \ell_i - \ell_{\text{base}}
\label{eq:marginal_contribution}
\end{equation}

A positive $\Delta_i$ indicates that repeating this frame increases the model's confidence in the correct answer, while a negative $\Delta_i$ suggests that repeating it introduces interference. 
This supervision signal comes directly from the MLLM's own feedback, requires no additional human annotation, and is tightly aligned with the downstream objective of answering the question correctly.

In practice, computing the repeat gain for all $N$ frames individually incurs substantial overhead. 
To mitigate this, we adopt a hybrid candidate sampling strategy: we form the candidate set $\mathcal{C}$ from the union of the top-$K$ frames ranked by the repeat scoring module's current scores and $K$ randomly sampled frames. 
The policy-ranked frames provide high-quality learning signals, while the random frames expand the exploration range and prevent the network from converging to local optima.

\subsection{Loss Design}
 Given the repeat gain $\{\Delta_i\}_{i \in \mathcal{C}}$ computed via AOI for the candidate set $\mathcal{C}$, we train the repeat scoring module to align its predicted scores with these supervision signals through two complementary losses.

\noindent\textbf{Regression Loss.} Since the raw scales of the predicted scores and repeat gain may differ significantly across samples, we first standardize both quantities within each sample:

\begin{equation}
\bar{s}_i = \frac{s_i - \mu_s}{\sigma_s + \epsilon}, \quad \bar{\Delta}_i = \frac{\Delta_i - \mu_\Delta}{\sigma_\Delta + \epsilon}
\label{eq:standardization}
\end{equation}

\noindent where $\mu_s, \sigma_s$ and $\mu_\Delta, \sigma_\Delta$ are the mean and standard deviation of the scores and repeat gain over the candidate set, respectively, and $\epsilon$ is a small constant for numerical stability. The regression loss is then defined as:

\begin{equation}
\mathcal{L}_{\text{reg}} = \frac{1}{|\mathcal{C}|} \sum_{i \in \mathcal{C}} \left(\bar{s}_i - \bar{\Delta}_i\right)^2
\label{eq:regression_loss}
\end{equation}

This formulation encourages the network to produce scores that preserve the relative ordering and magnitude of the repeat gain.

\noindent\textbf{Ranking Loss.} While the regression loss aligns scores with repeat gains in a point-wise manner, we additionally introduce a margin-based ranking loss to explicitly enforce the correct ordering between beneficial and detrimental frames. Specifically, we partition the candidate set into positive frames ($\mathcal{C}^+ = \{i \in \mathcal{C} \mid \Delta_i > 0\}$) and negative frames ($\mathcal{C}^- = \{i \in \mathcal{C} \mid \Delta_i < 0\}$), and require that positive frames are scored higher than negative frames by at least a margin $m$:

\begin{equation}
\mathcal{L}_{\text{rank}} = \frac{1}{|\mathcal{C}^+|} \sum_{i \in \mathcal{C}^+} \frac{1}{|\mathcal{C}^- \cup \mathcal{U}|} \sum_{j \in \mathcal{C}^- \cup \mathcal{U}} \max\!\left(0,\; s_j - s_i + m\right)
\label{eq:ranking_loss}
\end{equation}

\noindent where $\mathcal{U}$ denotes a randomly sampled subset of non-candidate frames, and $m$ is a margin hyperparameter. By including non-candidate frames as additional negatives, the ranking loss further encourages the network to assign higher scores to frames with positive repeat gain than to the remaining unevaluated frames.

\noindent\textbf{Total Loss.} The final training objective is a weighted combination of the two losses:

\begin{equation}
\mathcal{L} = \lambda_{\text{reg}} \cdot \mathcal{L}_{\text{reg}} + \lambda_{\text{rank}} \cdot \mathcal{L}_{\text{rank}}
\label{eq:total_loss}
\end{equation}

\noindent where $\lambda_{\text{reg}}$ and $\lambda_{\text{rank}}$ control the relative importance of the two objectives. 
The regression loss provides fine-grained score alignment with the repeat gain, while the ranking loss ensures that the critical qualitative distinction is robustly captured.
\section{Experiments}
\label{sec:exper}

\subsection{Evaluation Benchmarks}
We conduct a comprehensive evaluation of \methodname\ across five video benchmarks: VideoMME~\cite{fu2025video}, MLVU~\cite{zhou2024mlvu}, LongVideoBench~\cite{wu2024longvideobench}, LVBench~\cite{wang2025lvbench}, and Video-Holmes~\cite{cheng2025video}. 
These benchmarks are selected to cover a broad spectrum of video reasoning capabilities. 
VideoMME comprises videos spanning diverse themes and categorized into short, medium, and long durations, with each video paired with multiple questions to comprehensively assess performance. 
MLVU consists of seven subtasks designed to evaluate model capabilities from multiple distinct perspectives, including plot understanding, needle-in-a-haystack retrieval, egocentric reasoning, counting, ordering, anomaly recognition, and topic reasoning. 
LongVideoBench and LVBench focus on long-video understanding, featuring videos ranging up to several hours in length. 
Video-Holmes serves as a reasoning-intensive benchmark, specifically targeting complex multi-step video reasoning abilities.

\subsection{Implementation Details}
We train on top of Qwen2.5-VL-7B as the base model. 
The training data is derived from the multiple-choice subset of LLaVA-Video-178K~\cite{zhang2025llavavideovideoinstructiontuning} and the Video-Holmes~\cite{cheng2025video} training set. 
We perform data filtering by running inference with Qwen2.5-VL-7B five times per sample and discarding those that are answered either entirely correctly or entirely incorrectly across all five runs, resulting in approximately 20K training samples. 
During training, the video input is uniformly sampled at 128 frames. For the AOI supervision construction, both the top-$K$ and random-$K$ candidate sets use $K{=}16$, yielding 32 candidate frames per sample for training efficiency. The repeat scoring module contains only 3.7M trainable parameters, while the base VLM remains entirely frozen throughout training. 
We set the learning rate to $1\times10^{-4}$ with a per-device batch size of 1. 
The loss weights are $\lambda_{\text{reg}}{=}1.0$ and $\lambda_{\text{rank}}{=}0.1$. Training is conducted for one epoch using DeepSpeed ZeRO-2 with BF16 mixed precision.

\begin{table*}[t]
\centering

\setlength{\tabcolsep}{1.5mm}
\renewcommand{\arraystretch}{1.15}
\caption{\textbf{Main Results.} Comparison of five long-video understanding benchmarks. \imp{x} denotes improvement over the corresponding base model at the same frame count and * denotes results obtained under our settings. Best results per model group are in \textbf{bold}. We evaluate four frame repeating strategies: CLIP-K (based on CLIP scores), Bottom-K (lowest K scores), Random-K (random repeating), and FrameRepeat (Top-K scores).
LVB refers to LongVideoBench.}

\resizebox{\textwidth}{!}{
\begin{tabular}{l c ccccc}
\toprule
\textbf{Model} & \textbf{Frames} & \textbf{VideoMME} & \textbf{LVB} & \textbf{LVBench} & \textbf{MLVU} & \textbf{Video-Holmes} \\
\midrule

\rowcolor{lightgray}
\textcolor{graytext}{Qwen2.5-VL-72B} & \textcolor{graytext}{768} & \textcolor{graytext}{73.3} & \textcolor{graytext}{60.7} & \textcolor{graytext}{47.3} & \textcolor{graytext}{74.6} & \textcolor{graytext}{--} \\
\rowcolor{lightgray}
\textcolor{graytext}{Qwen3-VL-235B-A22B} & \textcolor{graytext}{768} & \textcolor{graytext}{79.0} & \textcolor{graytext}{--} & \textcolor{graytext}{63.6} & \textcolor{graytext}{83.8} & \textcolor{graytext}{--} \\
\rowcolor{lightgray}
\textcolor{graytext}{InternVL3.5-241B-A28B} & \textcolor{graytext}{--} & \textcolor{graytext}{72.9} & \textcolor{graytext}{67.1} & \textcolor{graytext}{--} & \textcolor{graytext}{78.2} & \textcolor{graytext}{--} \\

\midrule

\rowcolor{medgray}
\multicolumn{7}{c}{\textit{\small Qwen2.5-VL-7B-Instruct}} \\
\midrule

Base Model$^*$        & 64    & 59.6 & 48.0 & 32.9 & 57.1 & 33.4 \\
\quad + CLIP-K        & 64+4  & 60.2 & 49.6 & 33.8 & 59.4 & 33.0
\\
\quad + Bottom-K      & 64+4  & 59.6 & 48.9 & 32.3 & 55.9 & 33.4 \\
\quad + Random-K      & 64+4  & 60.3 & 48.7 & 33.2 & 56.8 & 33.7 \\
\rowcolor{greenhighlight}
\quad + \methodname~(Ours) & 64+4  & \textbf{61.0}\imp{1.4} & \textbf{50.8}\imp{2.8} & \textbf{34.3}\imp{1.4} & \textbf{60.3}\imp{3.2} & \textbf{34.9}\imp{1.5} \\
\midrule

Base Model$^*$        & 128   & 60.9 & 51.4 & 36.0 & 63.0 & 35.0 \\
\quad + CLIP-K        & 128+8 & 62.4 & 53.2 & 37.5 & 64.3  & 35.1
\\
\quad + Bottom-K      & 128+8 & 61.3 & 52.8 & 35.4 & 62.3 & 35.1 \\
\quad + Random-K      & 128+8 & 61.5 & 53.0 & 36.0 & 61.5 & 35.0 \\
\rowcolor{greenhighlight}
\quad + \methodname~(Ours) & 128+8 & \textbf{63.4}\imp{2.5} & \textbf{55.1}\imp{3.7} & \textbf{38.2}\imp{2.2} & \textbf{66.0}\imp{3.0} & \textbf{36.0}\imp{1.0} \\

\midrule

\rowcolor{medgray}
\multicolumn{7}{c}{\textit{\small Qwen3-VL-8B-Thinking}} \\
\midrule

Base Model$^*$        & 64    & 62.4 & 59.2 & 37.7 & 62.6 & 40.0 \\
\quad + CLIP-K        & 64+4 & 62.6 & 60.3 & 38.4 & 63.7& 40.2
\\
\quad + Bottom-K      & 64+4  & 62.0 & 58.5 & 36.7 & 63.9 & 39.3 \\
\quad + Random-K      & 64+4  & 62.3 & 58.9 & 38.1 & 64.5 & 39.3 \\
\rowcolor{greenhighlight}
\quad + \methodname~(Ours) & 64+4  & \textbf{63.2}\imp{0.8} & \textbf{61.5}\imp{2.3} & \textbf{39.3}\imp{1.6} & \textbf{65.4}\imp{2.8} & \textbf{40.5}\imp{0.5} \\
\midrule

Base Model$^*$        & 128   & 67.0 & 62.1 & 40.6 & 68.2 & 41.0 \\
\quad + CLIP-K        & 128+8 & 67.0 & 64.0 & 42.3 & 69.4 & 41.5 
\\
\quad + Bottom-K      & 128+8 & 66.3 & 63.5 & 41.2 & 68.3 & 41.0 \\
\quad + Random-K      & 128+8 & 67.3 & 63.3 & 41.6 & 68.5 & 40.2 \\
\rowcolor{greenhighlight}
\quad + \methodname~(Ours) & 128+8 & \textbf{67.9}\imp{0.9} & \textbf{65.4}\imp{3.3} & \textbf{44.3}\imp{3.7} & \textbf{71.2}\imp{3.0} & \textbf{43.4}\imp{2.4} \\

\bottomrule
\end{tabular}}
\label{tab:main_res}
\end{table*}

\subsection{Experimental Results}
\cref{tab:main_res} presents the experimental results of \methodname\ on five video understanding and reasoning benchmarks. 
We evaluate on two backbone models, Qwen2.5-VL-7B-Instruct and Qwen3-VL-8B-Thinking, and compare against three baseline frame repeating strategies: CLIP-K, Bottom-K and Random-K.

Across both models, \methodname\ achieves the best results on all benchmarks.
Taking the 128+8 frame setting on Qwen2.5-VL-7B as an example, our method yields improvements of +2.5, +3.7, +2.2, +3.0, and +1.0 on VideoMME, LongVideoBench, LVBench, MLVU, and Video-Holmes, respectively. 
On Qwen3-VL-8B-Thinking, the improvements are equally significant.
In contrast, Bottom-K and Random-K show limited gains and even degrade performance in certain settings, indicating that repeating low-importance or random frames may instead introduce interference. 
Although the CLIP-K sampling strategy achieves improvements on most benchmarks, it still falls short of our method, demonstrating the effectiveness of our AOI training.
The substantial advantage of \methodname\ over these three strategies demonstrates that our repeat scoring module can accurately predict the importance ranking of each frame, thereby effectively identifying and repeating the frames that truly contribute positively to the model's reasoning.
Furthermore, our repeat scoring module is trained only on Qwen2.5-VL-7B, yet achieves consistent improvements on Qwen3-VL-8B-Thinking as well, fully validating the cross-model generalization capability of \methodname.

% These results confirm that \methodname\ effectively mitigates visual forgetting by reinforcing key visual signals at the input stage, enabling the model to sustain attention to critical evidence throughout extended reasoning rather than drifting toward self-generated textual context.

\begin{table*}[t]
\centering
\begin{minipage}[t]{0.48\textwidth}
\centering
\small
\setlength{\tabcolsep}{2mm}
\renewcommand{\arraystretch}{1.15}
\caption{\textbf{Effect of Ranking Loss.} Performance with and without $\mathcal{L}_{\text{rank}}$ (128+8 frames).}

\resizebox{\linewidth}{!}{
\begin{tabular}{l l cc}
\toprule
\textbf{Model} & \textbf{Setting} & \textbf{VideoMME} & \textbf{Video-Holmes} \\
\midrule
\multirow{2}{*}{Qwen2.5-VL}
  & w/o $\mathcal{L}_{\text{rank}}$ & 62.1 & 35.5 \\
  & Full model & \textbf{63.4} & \textbf{36.0} \\
\midrule
\multirow{2}{*}{Qwen3-VL}
  & w/o $\mathcal{L}_{\text{rank}}$ & 66.8 & 42.1 \\
  & Full model & \textbf{67.9} & \textbf{43.4} \\
\bottomrule
\end{tabular}}
\label{tab:ablation_rank}
\end{minipage}
\hfill
\begin{minipage}[t]{0.48\textwidth}
\centering
\small
\setlength{\tabcolsep}{2mm}
\renewcommand{\arraystretch}{1.15}
\caption{\textbf{Comparison with using T\-SPO for frame repetition.} We use Qwen2.5-VL-7B as the base model and the 128+8 frame setting for tests.}

\resizebox{\linewidth}{!}{
\begin{tabular}{l cccc}
\toprule
\textbf{Model} & \textbf{VideoMME} & \textbf{LVB} & \textbf{LVBench}\\
\midrule
TSPO~\cite{tang2025tspo} & 62.8 & 54.3 & 36.9 \\
\methodname\   & \textbf{63.4} & \textbf{55.1} & \textbf{38.2} \\
\bottomrule
\end{tabular}}
\label{tab:ablation_select}
\end{minipage}
\end{table*}

\begin{table}[t]
\centering
\small
\setlength{\tabcolsep}{2.5mm}
\renewcommand{\arraystretch}{1.15}
\caption{\textbf{Effect of Repetition Count.} Comparison of repeating $K \in \{4, 8, 16\}$ frames on all benchmarks (128+$K$ frames). LVB refers to LongVideoBench.}

\resizebox{\columnwidth}{!}{
\begin{tabular}{l c ccccc}
\toprule
\textbf{Model} & $K$ & \textbf{VideoMME} & \textbf{LVB} & \textbf{LVBench} & \textbf{MLVU} & \textbf{Video-Holmes} \\
\midrule
\multirow{3}{*}{Qwen2.5-VL}
  & 4  & 62.3 & 53.9 & 36.6 & 64.7 & 35.2 \\
  & 8  & \textbf{63.4} & \textbf{55.1} & \textbf{38.2} & \textbf{66.0} & \textbf{36.0} \\
  & 16 & 62.9 & 53.2 & 36.0 & 65.6 & 34.6 \\
\midrule
\multirow{3}{*}{Qwen3-VL}
  & 4  & 67.3 & 62.8 & 41.7 & 70.0 & 41.4 \\
  & 8  & \textbf{67.9} & \textbf{65.4} & \textbf{44.3} & 71.2 & \textbf{43.4} \\
  & 16 & 67.3 & 65.0 & 43.7 & \textbf{71.3} & 43.0 \\
\bottomrule
\end{tabular}}
\label{tab:ablation_k}
\end{table}

\subsection{Ablation Study}
We conduct a series of ablation studies to validate the key design choices in \methodname.
Specifically, we examine four aspects: (1) the effect of the repetition count $K$, (2) the contribution of the ranking loss $\mathcal{L}_{\text{rank}}$, (3) the advantage of \methodname\ over frame selection methods, and (4) visualization of repeated frames.
Unless otherwise stated, we adopt the same training configuration as the full model.

\paragraph{\textbf{Effect of Repetition Count.}} 
We investigate the impact of varying the number of repeated frames $K$. 
As shown in~\cref{tab:ablation_k}, we compare $K \in \{4, 8, 16\}$ on all five benchmarks. 
On both models, increasing $K$ from 4 to 8 yields consistent improvements across nearly all benchmarks. 
However, further increasing to $K\!=\!16$ shows mixed results: on Qwen2.5-VL, performance drops on all benchmarks compared to $K\!=\!8$; on Qwen3-VL, $K\!=\!16$ surpasses $K\!=\!8$ on MLVU, but declines on the other benchmarks. 
These results suggest that $K\!=\!8$ provides the best overall trade-off, as excessive repetition may introduce redundancy that dilutes the reinforcement effect.

\paragraph{\textbf{Effect of Ranking Loss.}}
We ablate the ranking loss $\mathcal{L}_{\text{rank}}$ by training with only the regression loss $\mathcal{L}_{\text{reg}}$.
As shown in~\cref{tab:ablation_rank}, removing $\mathcal{L}_{\text{rank}}$ leads to consistent performance drops on both models: VideoMME decreases by 1.3 and 1.1 points on Qwen2.5-VL and Qwen3-VL, respectively, and Video-Holmes drops by 0.5 and 1.3 points. 
This confirms that the ranking loss provides complementary pairwise supervision that helps the network better distinguish frame importance orderings beyond pointwise regression alone.

\paragraph{\textbf{Comparison with Frame Selection Methods.}}

To further validate our approach, we compare it against a baseline that directly repurposes TSPO~\cite{tang2025tspo}, a text and event aware frame selection method, for frame repetition. 
Although frame selection and frame repetition appear related, they serve fundamentally different objectives: the former identifies the most informative frames to \textbf{retain}, while the latter seeks to reinforce \textbf{temporally critical yet underattended cues}. 
Simply selecting the highest-scored frames for repetition tends to over-emphasize already salient content, rather than compensating for the model's temporal blind spots. 
Results across multiple benchmarks shown in~\cref{tab:ablation_select} consistently demonstrate that our method outperforms the TSPO-based repetition baseline, confirming that a learned repetition strategy is superior to naively reusing scores from a frame selection paradigm.

\begin{figure*}[t]
\centering
\includegraphics[width=\linewidth]{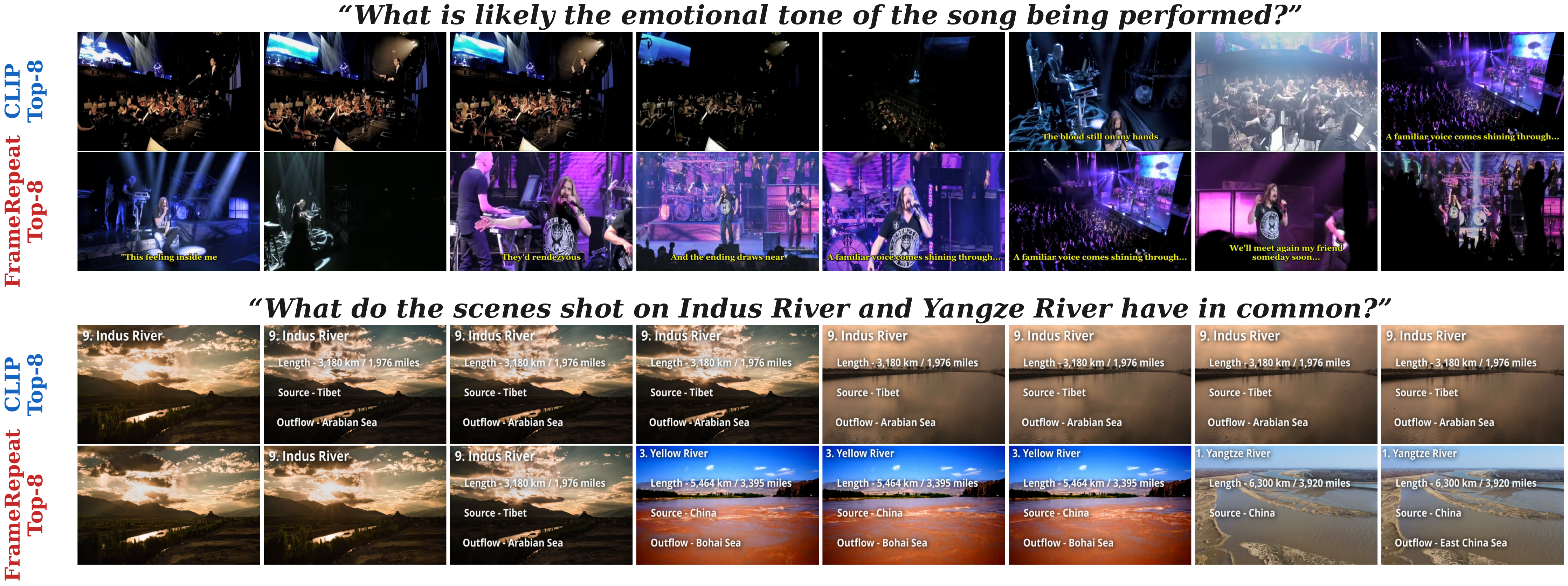}
\caption{Comparison of frames selected using \methodname\ score versus CLIP score.}
\label{fig:visualization}
\end{figure*}

\paragraph{\textbf{Visualization.}}
We visualize the frames selected for repetition by \methodname\ and compare them with those selected using CLIP score. 
The results are shown in~\cref{fig:visualization}. 
As can be seen, CLIP score selects frames based solely on global feature similarity, which makes it difficult to capture fine-grained semantic information relevant to the question. In contrast, our frame repeat module, after being trained with AOI, can more accurately assess the importance of each frame for answering the question by incorporating question semantics, thereby selecting frames that truly contain key visual clues. 
By repeating these frames, the model is better guided to attend to the important information during its reasoning process, ultimately correcting previously incorrect predictions.
\section{Conclusion}
\label{sec:conclusion}
In this work, we propose \methodname\ to mitigate the dilution of visual signals during video reasoning. 
By repeating key frames at the input stage, we amplify the visual signal strength of these frames, enabling the model to sustain attention to critical information throughout long chain-of-thought reasoning. 
We design a lightweight repeat scoring module to predict each frame's impact on producing the correct answer, trained via an Add-One-In (AOI) strategy that learns the marginal information contribution of individual frames. 
Our experimental results demonstrate that \methodname\ effectively improves model performance on video reasoning tasks and exhibits strong generalization ability, validating the effectiveness of frame repetition in alleviating visual signal dilution during the thinking process.

% ---- Bibliography ----
%
% BibTeX users should specify bibliography style 'splncs04'.
% References will then be sorted and formatted in the correct style.
%
\bibliographystyle{splncs04}
\bibliography{main}
\appendix
\clearpage
\setcounter{page}{1}

\chapter*{Supplementary Material for When Thinking Hurts: Mitigating Visual Forgetting in Video Reasoning via Frame Repetition}

\section{Implementation Details}

The detailed training hyperparameters of FrameRepeat are summarized in~\cref{tab:hyperparams}. 
We train the repeat scoring module on top of frozen CLIP-ViT-Large~\cite{radford2021learning} visual and textual features, using a combination of regression loss and ranking loss to supervise frame importance scoring. 
The training is conducted with DeepSpeed ZeRO~\cite{rajbhandari2020zero} Stage 2 and mixed-precision (BF16) for efficiency. 
We adopt a moderate learning rate with gradient accumulation to maintain a stable effective batch size throughout training.

\begin{table}[h]
\centering
\caption{Training hyperparameters of FrameRepeat.}
\label{tab:hyperparams}
\begin{tabular}{ll}
\toprule
\textbf{Hyperparameter} & \textbf{Value} \\
\midrule
\multicolumn{2}{l}{\textit{Model}} \\
Base Model & Qwen2.5-VL-7B \\
CLIP Model & CLIP-ViT-Large \\
Parameters & 3.69M \\
Hidden Dim / Heads & 768 / 8 \\
CLIP Prior Weight $\lambda$ & 5.0 \\
\midrule
\multicolumn{2}{l}{\textit{Frame Sampling}} \\
Base Frames & 128 \\
Top-$k$ (repeat) & 16 \\
$|\mathcal{C}|$ & 32 (Top-16 + Random-16) \\
\midrule
\multicolumn{2}{l}{\textit{Training}} \\
DeepSpeed & ZeRO Stage 2 \\
Precision & BF16 \\
Learning Rate & $1 \times 10^{-4}$ \\
Epochs & 1 \\
Per-device Batch Size & 1 \\
Gradient Accumulation Steps & 4 \\
Effective Batch Size & 32 \\
Num Generations & 16 \\
Max Completion Length & 256 \\
\midrule
\multicolumn{2}{l}{\textit{Loss Weights}} \\
Regression Weight $\lambda_{\text{reg}}$ & 1.0 \\
Ranking Weight $\lambda_{\text{rank}}$ & 0.1 \\
\bottomrule
\end{tabular}
\end{table}

\section{Further Experiments}
\subsection{Tests with Repeated Frames Only}

To validate the effectiveness of our frame repetition strategy, we conduct an ablation study where only the top-k frames selected by FrameRepeat are fed into the model, while the remaining frames are entirely discarded. 
As shown in~\cref{tab:select_only_videomme}, this approach leads to a substantial performance degradation across all settings compared to the full pipeline that retains all original frames alongside the repeated ones.
Even as the number of selected frames increases, the accuracy remains significantly lower. 
This demonstrates that the non-repeated frames play an essential role in providing complementary temporal context — simply removing them deprives the model of the broader scene understanding necessary for accurate video comprehension. 
In contrast, our repetition-based approach effectively amplifies the influence of key frames while preserving the complete temporal context, enabling the model to achieve superior performance without sacrificing any visual information.

\begin{table}[h]
\centering
\caption{VideoMME~\cite{fu2025video} accuracy (\%) under select-only mode with varying top-$k$ frames.
We first uniformly sample 128 frames, then sample k frames from the 128 frames for model input.}
\label{tab:select_only_videomme}
\begin{tabular}{lcccc}
\toprule
\textbf{Model} & \textbf{Top-8} & \textbf{Top-16} & \textbf{Top-32} & \textbf{Top-64} \\
\midrule
Qwen2.5-VL-7B~\cite{bai2025qwen25vltechnicalreport}  & 48.3 & 52.1 & 54.6 & 58.0 \\
Qwen3-VL-8B-Thinking~\cite{bai2025qwen3vltechnicalreport}   & 52.1 & 54.6 & 57.9 & 62.5 \\
\bottomrule
\end{tabular}
\end{table}

\begin{table}[h]
\centering
\caption{Comparison between Qwen2.5-VL base (128 frames) and with FrameRepeat (top-8 repetition) under no-think mode.}
\label{tab:base_vs_repeat_no_think}
\begin{tabular}{lccccc}
\toprule
\textbf{Method} & \textbf{VideoMME} & \textbf{MLVU} & \textbf{LVB} & \textbf{LVBench} & \textbf{VideoHolmes} \\
\midrule
Base (128f)    & 65.1 & 69.3 & 61.2 & 40.7 & 41.6 \\
+ FrameRepeat  & 66.6 & 69.4 & 61.6 & 43.4 & 41.5 \\
\bottomrule
\end{tabular}
\end{table}

\subsection{Effect of FrameRepeat on No Thinking Settings}

To investigate whether frame repetition can effectively reinforce key visual information at the input level, we evaluate FrameRepeat under the no-thinking setting, where the model directly outputs an answer without chain-of-thought reasoning. 
As shown in~\cref{tab:base_vs_repeat_no_think}, FrameRepeat consistently improves performance across the majority of benchmarks even without explicit reasoning. 
Since the no-thinking setting eliminates the model's ability to iteratively reason over the input, any performance gain can be more directly attributed to the enhanced saliency of critical visual content introduced by frame repetition. 
This confirms that repeating key frames serves as an effective mechanism for reinforcing question-relevant visual information, making it more prominent during the model's encoding stage and thus easier to leverage for accurate prediction.

\subsection{Discussion of Relationship with Frame Selection Methods}
Frame selection and frame repetition share a conceptual similarity: both aim to identify critical visual information and enhance it. 
However, they operate through complementary mechanisms. 
Frame selection improves video understanding by removing distracting or redundant frames, which inherently requires retaining a larger number of frames to preserve temporal coherence. 
In contrast, frame repetition strengthens key visual information by duplicating a small number of critical frames, without discarding any part of the original sequence.

We further investigate whether frame selection methods can be directly repurposed for frame repetition. 
As shown in the main experiments, using a frame selection model (TSPO~\cite{tang2025tspo}) to choose frames for repetition yields inferior results compared to our dedicated FrameRepeat model. 
This indicates that simply reinforcing the frames deemed most important by a selection model is insufficient — effective frame repetition requires targeted training that teaches the model which frames, when duplicated, most benefit downstream VLM reasoning.

Moreover, we demonstrate that frame repetition and frame selection are not mutually exclusive but can be effectively combined. 
As shown in~\cref{tab:tspo_repeat}, we first apply TSPO to select 64 frames from the original video, then employ FrameRepeat to identify and duplicate additional key frames from this reduced set. 
The VLM is prompted to generate chain-of-thought reasoning before producing the final answer. 
Results across three benchmarks show that FrameRepeat consistently improves upon the already-filtered frame set, with accuracy gains on VideoMME (+1.6\%), and LongVideoBench (+1.4\%). 
This confirms that the two approaches address orthogonal aspects of temporal understanding and can be synergistically combined: frame selection provides a cleaner temporal signal by removing noise, while frame repetition further amplifies the most informative visual cues within the selected frames.

\begin{table}[t]
\centering
\caption{Comparison of TSPO 64 frames and TSPO 64 frames + FrameRepeat on three benchmarks.}
\label{tab:tspo_repeat}
\begin{tabular}{lccc}
\toprule
\textbf{Method} & \textbf{VideoMME} & \textbf{MLVU} & \textbf{LongVideoBench} \\
\midrule
TSPO 64 frames              & 59.6 & 71.2 & 53.2 \\
+ Repeat Top-4        & \textbf{61.2} & \textbf{72.1} & \textbf{54.6} \\
+ Repeat Top-8        & 60.9 & 71.2 & 53.4 \\

\bottomrule
\end{tabular}
\end{table}

\section{Training Dynamics}

\begin{figure}[t]
    \centering
    \includegraphics[width=\linewidth]{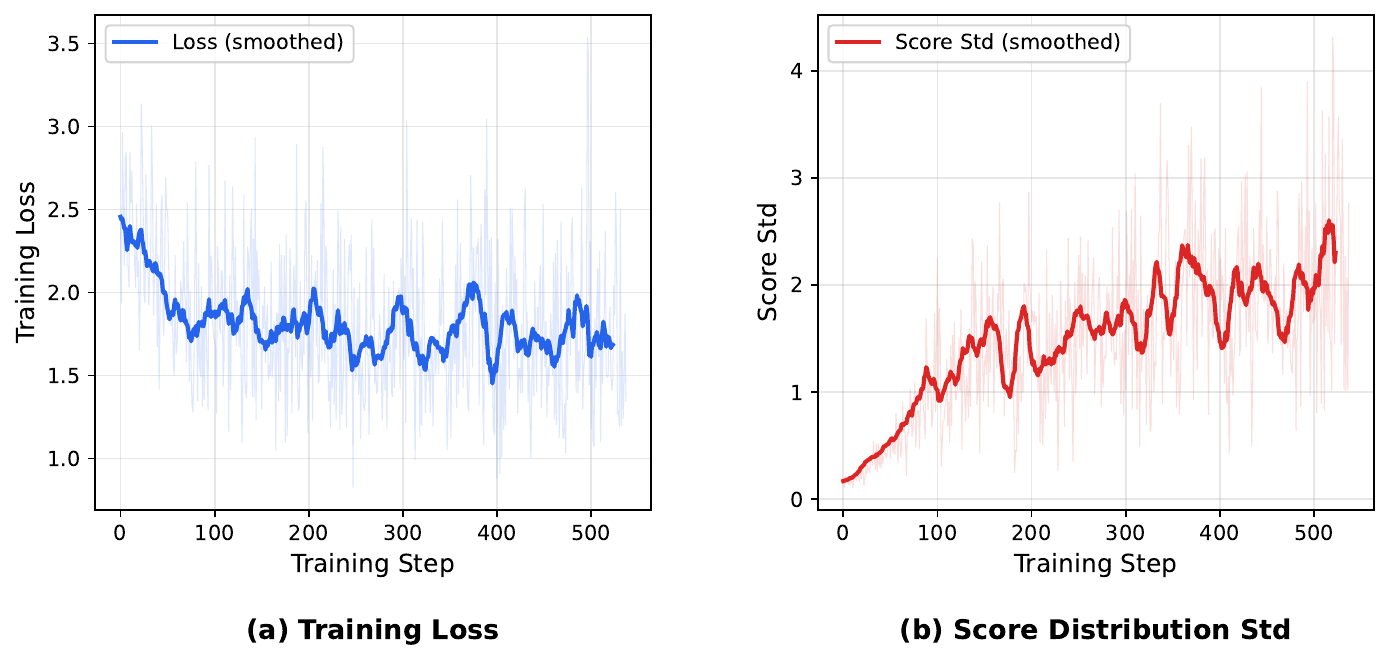}
    \caption{\textbf{Training dynamics of the repeat scoring module.} (a) The training loss decreases steadily, indicating stable convergence. (b) The score distribution standard deviation increases consistently, demonstrating that the model progressively learns to differentiate key frames from non-essential ones.}
    \label{fig:training_curves}
\end{figure}

\begin{figure*}[t]
\centering
\includegraphics[width=\textwidth]{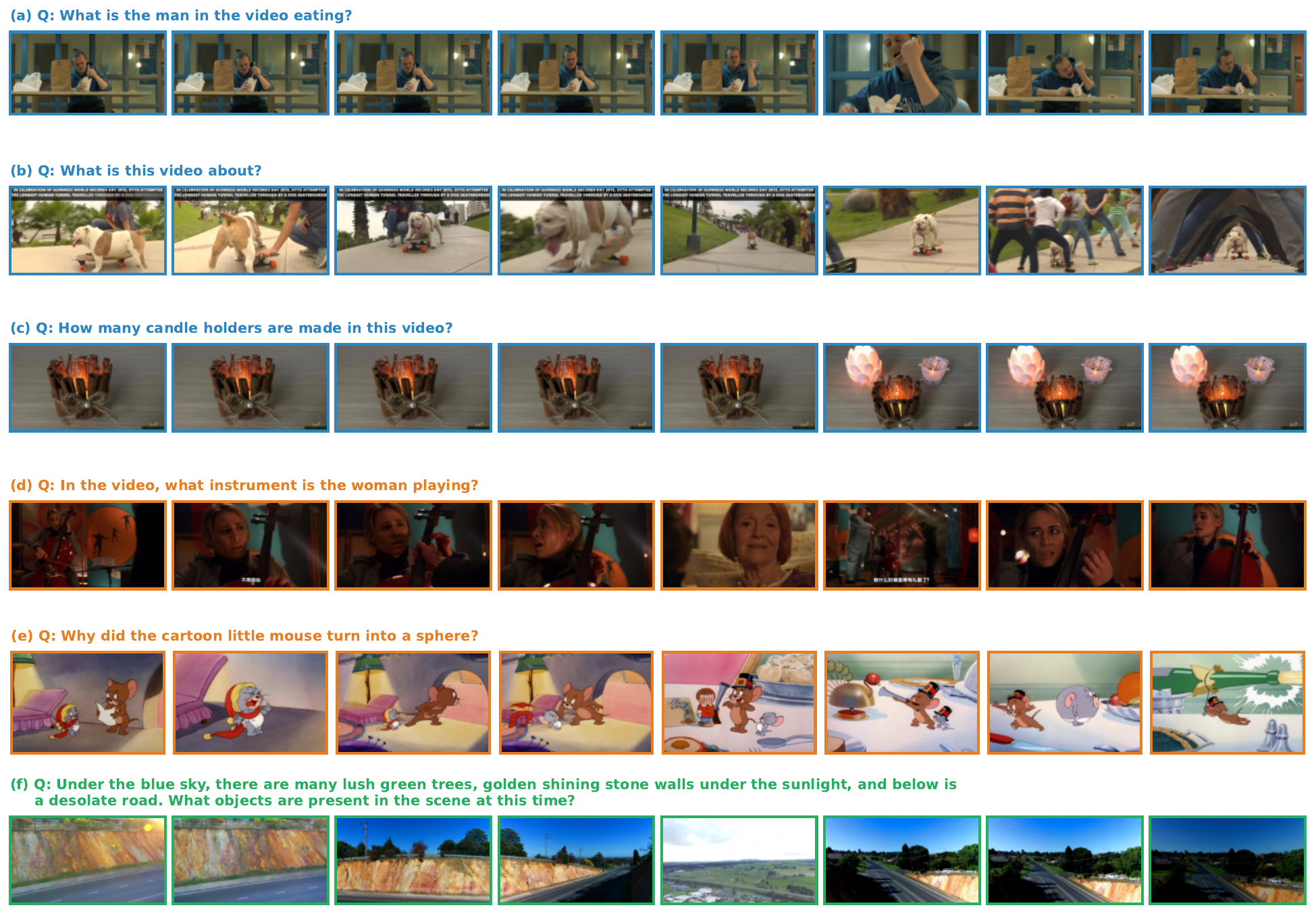}
\caption{Visualization of the top-8 frames selected by FrameRepeat for repetition across six examples from VideoMME~\cite{fu2025video}, MLVU~\cite{zhou2024mlvu}, and LongVideoBench~\cite{wu2024longvideobench}. Each row shows the repeated frames for a given question. }
\label{fig:vis}
\end{figure*}

\begin{figure*}[t]
\centering
\includegraphics[width=\textwidth]{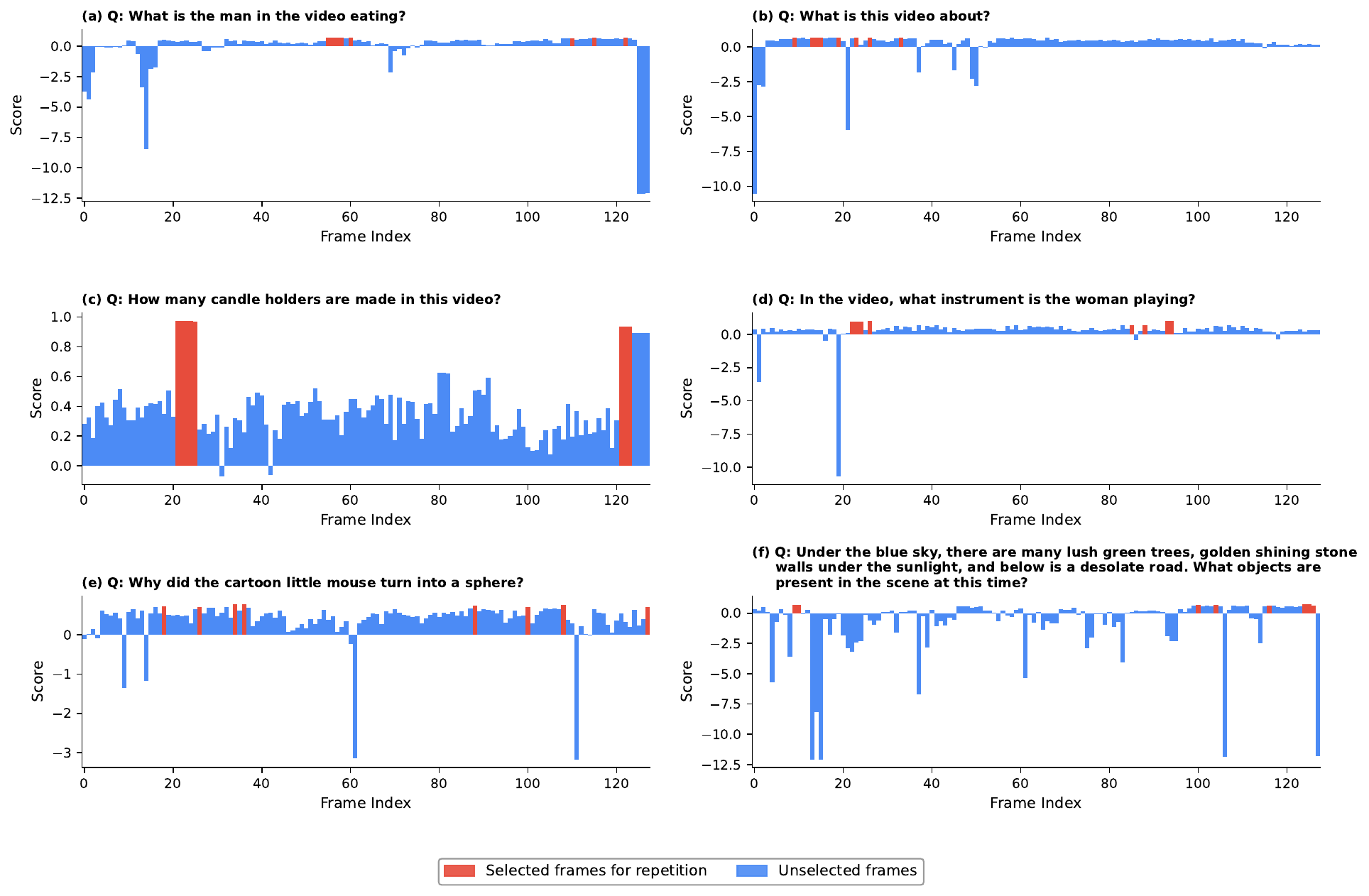}
\caption{Frame importance score distributions predicted by FrameRepeat for the six examples in ~\cref{fig:vis}. Red bars indicate the frames selected for repetition. The selected frames consistently correspond to the highest-scoring positions, and their distribution across the temporal axis reflects the model's ability to identify question-relevant moments throughout the video.}
\label{fig:score_distribution}
\end{figure*}

\cref{fig:training_curves} illustrates the training dynamics of the repeat scoring module. 
As shown in~\cref{fig:training_curves}(a), the training loss decreases steadily throughout the optimization process, indicating stable convergence of the joint regression and ranking objectives. 
More notably,~\cref{fig:training_curves}(b) reveals a consistent increase in the standard deviation of predicted frame scores over the course of training.
This trend demonstrates that the policy model progressively learns to differentiate between key frames and non-essential ones, transitioning from initially assigning near-uniform scores to all frames toward producing increasingly discriminative score distributions. 
The growing score dispersion reflects the model's enhanced ability to identify question-relevant visual content, which serves as the foundation for effective frame repetition.

\section{Visualization}

We visualize the top-8 frames selected by FrameRepeat for six representative examples in ~\cref{fig:vis}. 
The results show that the model consistently identifies frames containing the visual content most relevant to the given question, and the selected frames are temporally distributed across the video rather than clustered in a single segment, confirming that FrameRepeat learns a meaningful and question-aware frame importance signal.

We further visualize the frame importance score distributions in ~\cref{fig:score_distribution}.
The frames selected are highlighted in red.

\section{Prompt}

\begin{tcolorbox}[
  title=\textbf{Training Prompt Template (AOI Scoring)},
  colback=blue!3!white,
  colframe=blue!50!black,
  fonttitle=\bfseries,
  breakable
]
\texttt{\{Question\}} Please answer with the option's letter from the given choices directly.

Answer:~
\end{tcolorbox}

\begin{tcolorbox}[
  title=\textbf{Evaluation Prompt Template (Think)},
  colback=green!3!white,
  colframe=green!50!black,
  fonttitle=\bfseries,
  breakable
]
\texttt{\{Question\}} Only give the best option. You FIRST think about the reasoning process as an internal monologue and then provide the final answer. The reasoning process MUST BE enclosed within \texttt{<think>}\texttt{</think>} tags. The final answer MUST BE put within \texttt{<answer>}\texttt{</answer>} tags.
\end{tcolorbox}

\begin{tcolorbox}[
  title=\textbf{Evaluation Prompt Template (No-Think)},
  colback=orange!3!white,
  colframe=orange!50!black,
  fonttitle=\bfseries,
  breakable
]
\texttt{\{Question\}} Only give the best option.
\end{tcolorbox}
\end{document}